%% file: main.tex
\documentclass[10pt,twocolumn,letterpaper]{article}

\usepackage[pagenumbers]{cvpr} 
\usepackage{multirow}
\usepackage{caption}
\usepackage[accsupp]{axessibility}  

\makeatletter
\def\blfootnote{\xdef\@thefnmark{}\@footnotetext}
\makeatother

\input{preamble}
\definecolor{cvprblue}{rgb}{0.21,0.49,0.74}
\usepackage[pagebackref=false,breaklinks,colorlinks,allcolors=cvprblue]{hyperref}

\def\paperID{10} 
\def\confName{CVPR}
\def\confYear{2026}
\newcommand{\formattedparagraph}[1]{\noindent \textbf{#1}}

\title{Rethinking Dense Optical Flow without Test-Time Scaling}

\author{\quad Praroop Chanda$^{1, 3}$\quad Suryansh Kumar$^{1, 2, 3, 4, *}$\\
Visual and Spatial AI Lab${^1}$, VCCM Section  \\ College of PVFA${^2}$, Department of ECEN$^3$, Department of CSCE$^4$, \\ Texas A\&M University, College Station, Texas, USA
}

\begin{document}
\maketitle
\input{sec/0_abstract}
\input{sec/1_intro}

\input{sec/2_related}
\input{sec/3_method}

\input{sec/4_experiments}

\input{sec/5_conclusion}
{
    \small
    \bibliographystyle{ieeenat_fullname}
    \bibliography{main}
}

\end{document}

%% file: sec/0_abstract.tex
\begin{abstract}
Recent progress in dense optical flow has been driven by increasingly complex architectures and multi-step refinement for test-time scaling. While these approaches achieve strong benchmark performance, they also require substantial computation during inference. This raises a fundamental question: Is scaling test-time computation the only way to improve dense optical flow accuracy? We argue that it is not. Instead, powerful visual semantic and geometric priors encoded in modern foundation models can reduce, if not overcome, the need for computationally expensive iterative refinement at test-time. In this paper, we present a framework that estimates dense optical flow in a single forward pass, leveraging pretrained foundation representations, while avoiding iterative refinement and additional inference-time computation, thus offering an alternative to test-time scaling. Our method extracts visual semantic features from a frozen DINO-v2 backbone and combines them with geometric cues from a monocular depth foundation model. We fuse these complementary priors into a unified representation and apply a global matching formulation to estimate dense correspondences without recurrent updates or test-time optimization. Despite avoiding iterative refinement, our approach achieves strong cross-dataset generalization across challenging benchmarks. On Sintel Final, we obtain \textbf{2.81 EPE} without refinement, significantly improving over state-of-the-art (SOTA) SEA-RAFT under comparable training conditions and outperforming RAFT, GMFlow (without refinement), and recent FlowSeek in the same setting. These results suggest that strong foundation priors can substitute for test-time scaling, offering a computationally efficient alternative to refinement-heavy pipelines.
\end{abstract}

%% file: sec/1_intro.tex
\section{Introduction}
\label{sec:intro}
Dense optical flow is a fundamental problem in computer vision, robotic perception, and machine vision\blfootnote{*Corresponding Author: Suryansh Kumar}. The task seeks to estimate a optical flow vector for every pixel between two consecutive images \citep{sun2010secrets}. Accurate estimation of optical flow enables a wide range of downstream applications, including video understanding \citep{gao2020flow}, saliency detection \citep{li2019motion}, action recognition \citep{piergiovanni2019representation, sun2018optical}, dense 3D reconstruction \citep{qian2026flow4r, kumar2019superpixel, chen2023uncertainty, kumar2017monocular}, shape deformation modeling \cite{kumar2022organic, kumar2019jumping, kumar2020non, kumar2018scalable}, and motion modeling \citep{huang2024zero}. Over the past decade, deep learning methods have dramatically improved optical flow accuracy. Architectures such as RAFT \citep{teed2020raft} and its variants \citep{wang2024sea} achieve near-saturating performance on standard benchmarks. However, these improvements come at an increasing computational cost. Modern pipelines rely heavily on large annotated datasets, extensive training schedules, and most critically multi-step test-time refinement, which requires substantial inference-time compute.

This trend naturally raises a fundamental question: Is allocating more computation at time-time the only way forward to improve optical flow performance? Recent developments in foundation models for visual representation suggest an alternative viewpoint. Large-scale pretrained models trained once on internet-scale datasets have demonstrated strong transferability across diverse visual tasks, including object detection \citep{zhuang2025argus}, segmentation \citep{zhang2024improving}, depth estimation \citep{yang2024depth, yang2024depthv2}, and geometric reasoning \citep{yang2024depth, liu2023segment}. Rather than repeatedly training specialized architectures or increasing inference-time computation, these models show that powerful pretrained representations can generalize across domains and tasks. This observation raises an intriguing possibility for dense optical flow, i.e.,  instead of relying on increasingly expensive test-time refinement unlike SOTA \cite{wang2024sea, Poggi_2025_ICCV}, we can leverage strong visual semantic and geometric priors from foundation models to perform accurate motion estimation directly.

Surprisingly, dense optical flow has remained largely disconnected from such paradigm despite the rapid progress of foundation models in other vision domains. Current state-of-the-art approaches \citep{wang2024sea, teed2020raft, Poggi_2025_ICCV} continue to emphasize architectural innovations, dataset scaling, and iterative update mechanisms, treating optical flow as a task that must be explicitly learned and repeatedly corrected during inference. While these strategies improve benchmark performance, they also reinforce the assumption that accurate optical flow estimation requires task-specific feature encoders and expensive refinement procedures. Nevertheless, as we know, dense optical flow is fundamentally a correspondence problem, where the key challenge is to learn representations that reliably match pixels across frames while respecting scene structure and motion boundaries. Modern vision foundation models already encode many of these properties, including strong semantic discrimination and boundary-aware geometric cues. This observation suggests that dense optical flow may benefit more from reusing powerful pretrained representations than from further increasing architectural complexity or test-time computation.

In this work, we revisit the dense optical flow problem from a foundation model perspective. We argue that strong semantic and geometric priors contained in modern pretrained models can significantly reduce the reliance on test-time scaling through iterative refinement. Instead of allocating additional computation during inference, we investigate whether carefully designed representations can enable accurate motion estimation in a single forward pass. Our central hypothesis is that dense optical flow can emerge from sufficiently powerful visual semantic and geometric representations without repeated correction through iterative updates.

Concretely, we introduce a dense optical flow framework that leverages two complementary foundation priors. First, we extract semantic visual features from the self-supervised DINO-v2 model \citep{oquab2023dinov2}, which provides spatially coherent representations capable of capturing fine-grained visual correspondence. Second, we incorporate geometric structure using a monocular depth foundation model \citep{yang2024depthv2}. Although depth prediction is learned from data rather than derived from explicit geometry, modern depth prediction models produce high-frequency structural cues and sharp boundaries that are particularly informative for correspondence estimation near occlusions and motion discontinuities. By combining these semantic and geometric signals, we construct a unified representation suitable for dense matching.

The proposed framework fuses DINO-v2 features and depth foundation representations within a global matching formulation inspired by GMFlow \citep{xu2022gmflow}. Unlike RAFT \citep{teed2020raft}, SEA-RAFT \citep{wang2024sea}, and very recently proposed FlowSeek \citep{Poggi_2025_ICCV}, our model performs optical flow estimation in a single forward pass and does not require iterative refinement to achieve competitive performance. Empirically, our approach demonstrates strong cross-dataset generalization across challenging benchmarks. On the Sintel Final benchmark, we achieve 2.81 EPE without refinement, substantially improving over SEA-RAFT (4.32 EPE) under comparable training conditions and outperforming RAFT, GMFlow (without refinement), and FlowSeek in the same setting. These results suggest that strong foundation priors can partially substitute for test-time scaling in dense optical flow estimation. More broadly, our work highlights the potential of foundation-model-driven inference pipelines to reduce the reliance on refinement-heavy architectures and large task-specific training pipelines. Our \textbf{contributions} are summarized as follows:

\begin{itemize}
    \item We propose a dense optical flow estimation framework that operates without test-time refinement and leverages frozen pretrained foundation models instead of training task-specific encoders.
    \item We show that combining DINO-v2 visual semantic features \citep{oquab2023dinov2} with depth foundation representations \citep{yang2024depthv2} provides a powerful prior for dense optical flow correspondence estimation.
    \item Our work demonstrate that strong pretrained representations can reduce reliance on iterative inference procedures. Empirically, our framework achieves competitive and often superior performance compared to refinement-based SOTA approaches such as RAFT \citep{teed2020raft}, GMFlow \citep{xu2022gmflow}, SEA-RAFT \citep{wang2024sea}, and FlowSeek \citep{Poggi_2025_ICCV}, including \textbf{2.81 EPE} on Sintel Final without refinement.
\end{itemize}

%% file: sec/2_related.tex
\section{Related Work}
\label{sec:relatedwork}

\formattedparagraph{\textit{(i)} Learning-Based Optical Flow.}
Deep learning has greatly advanced the accuracy of dense optical flow estimation. Early convolutional architectures such as FlowNet \citep{dosovitskiy2015flownet, ilg2017flownet} and PWC-Net \citep{sun2018pwc} positioned optical flow as a supervised regression problem over learned cost volumes, while subsequent approaches introduced multi-scale processing and feature warping strategies to improve efficiency and robustness \citep{sun2019models, jahedi2024ms}. A major change occurred with RAFT \citep{teed2020raft}, which reformulated optical flow estimation as iterative refinement over an all-pairs correlation field. RAFT dramatically improved accuracy yet introduced recurrent update operators that require multiple refinement steps at test time. SEA-RAFT \citep{wang2024sea} further improved this paradigm by simplifying the architecture and proposing probabilistic supervision and rigid-motion pretraining, showing a strong balance between efficiency and accuracy. Despite these improvements, current SOTA pipelines still rely heavily on multi-step test-time iterative refinement to gain performance.

Transformer-based approaches have also been explored to improve global correspondence reasoning. Methods such as GMFlow \citep{xu2022gmflow} and FlowFormer \citep{huang2022flowformer} leverage attention mechanisms to capture long-range dependencies and handle large displacements. GMFlow reformulates optical flow estimation as a global matching problem using attention and softmax normalization, removing recurrent updates but still learning flow-specific feature representations through supervised training. FlowFormer extends this idea by proposing cost-volume transformers to better model long-range dependencies. While these models improve global reasoning, they depend on task-specific feature learning and often benefit from additional refinement.

\smallskip
\formattedparagraph{\textit{(ii)} Geometric Priors in Optical Flow.} Incorporating geometric cues into optical flow has long been explored as a way to improve correspondence estimation \citep{sun2010secrets}. Earlier work often relied on piecewise rigid motion assumptions or geometric regularization \citep{vogel2013piecewise}. More recently, learning-based approaches have begun integrating depth priors \cite{liuva, liu2023single} into optical flow pipelines. FlowSeek \citep{Poggi_2025_ICCV} represents a very recent example of this direction. It injects monocular depth predictions from foundation models \citep{yang2024depthv2} into a RAFT-style architecture and introduces motion bases to improve cross-dataset generalization. However, FlowSeek still trains a flow-specific backbone and relies heavily on iterative refinement to realize its performance gains. In contrast, our work treats depth predictions as representation priors rather than refinement guidance. Instead of embedding depth within a recurrent refinement pipeline, we directly fuse depth-aware features with semantic representations to enable single-pass correspondence estimation.

\smallskip
\formattedparagraph{\textit{(iii)} Foundation Models as Visual Representation.} Vision foundation models have recently demonstrated remarkable transferability across tasks. Self-supervised models such as DINO \citep{caron2021emerging} and DINO-v2 \citep{oquab2023dinov2} learn semantically meaningful feature representations that generalize well across domains without task-specific supervision. In parallel, depth foundation models such as Depth Anything \citep{yang2024depth, yang2024depthv2} have shown strong cross-dataset generalization while producing high-frequency geometric cues and sharp boundary predictions. Despite the success of foundation models across many visual tasks, their role in dense optical flow remains largely unexplored. Existing optical flow pipelines typically retain task-specific encoders even when incorporating geometric priors. In this work, we instead treat foundation models as frozen representation priors and focus on designing an inference pipeline that exploits their semantic and geometric cues. By combining DINO-v2 features with depth foundation representations and performing global correspondence estimation without iterative refinement, our approach aligns dense optical flow with the broader paradigm of foundation-driven inference, reducing reliance on task-specific training and test-time scaling.

Overall, our work differs from existing dense optical flow models in three key aspects. First, unlike refinement-based methods such as RAFT \citep{teed2020raft} and its extension SEA-RAFT \citep{wang2024sea}, which rely on recurrent update operators and multi-step test-time refinement, our framework estimates optical flow in a single forward pass without allocating additional inference-time computation. Second, although our matching formulation follows the global correspondence strategy introduced in GMFlow \citep{xu2022gmflow}, we depart from GMFlow by removing flow-specific feature learning and instead relying on frozen foundation representations to drive correspondence estimation. Third, while FlowSeek \citep{Poggi_2025_ICCV} integrates depth foundation models into a refinement-based architecture, it still trains a dedicated flow backbone and depends on iterative updates to achieve its performance gains. In contrast, we utilize both vision and geometric foundation models as fixed representation priors and focus on designing a lightweight inference pipeline that leverages them for dense optical flow correspondence estimation. Taken together, our approach shifts the emphasis from increasing architectural complexity or test-time scaling toward representation-driven optical flow, where strong pretrained semantic and geometric features enable accurate motion estimation without iterative refinement.

%% file: sec/3_method.tex
\section{Method}\label{sec:methodology}
Given two consecutive RGB frames $\mathbf{I}_1 \in \mathbb{R}^{H \times W \times 3}, \mathbf{I}_2 \in \mathbb{R}^{H \times W \times 3}$, our goal is to estimate dense optical flow $\mathbf{V} \in \mathbb{R}^{H \times W \times 2}$, where $H, W$ denote the height and width of the image, respectively.   We first extract dense visual features using a vision foundation model (Sec. \ref{ssec:featureextraction}), and then incorporate monocular depth priors to encode geometric structure (Sec. \ref{ssec:depthpriorfeature}). These complementary cues are integrated through a joint cross-modal feature fusion strategy (Sec. \ref{ssec:crossmodalfeature}), producing a unified representation for dense correspondence estimation. Finally, we apply a transformer-based global matching and flow propagation pipeline (Sec. \ref{ssec:transformerbasedglobalmatching}) to infer optical flow in a single-pass, without iterative refinement or test-time optimization. In contrast to refinement-based optical flow pipelines, our framework estimates dense correspondences in a single forward pass by leveraging pretrained semantic and geometric priors. The overall loss function is provided in Sec. \ref{ssec:trainingloss}.

\subsection{Foundation Visual Feature Extraction} \label{ssec:featureextraction}
Rather than learning a task-specific feature encoder as in prior optical flow methods \citep{wang2024sea, Poggi_2025_ICCV}, we reuse pretrained semantic representations from vision foundation models. We begin by extracting dense semantic visual representations from each input frame using a frozen vision foundation model \citep{oquab2023dinov2}. Specifically, given $\mathbf{I}_1, \mathbf{I}_2$, we compute feature embeddings via a pretrained DINOv2-S (small) \citep{oquab2023dinov2} encoder:
\begin{equation}
\mathbf{F}_i^{D} = \Phi_{\text{DINO}}(\mathbf{I}_i), 
\quad i \in \{1,2\},
\end{equation}
where, $\Phi_{\text{DINO}}$ denotes the DINOv2 backbone trained with large-scale self-supervised learning. The resulting feature maps are extracted at a spatial resolution of $\frac{1}{8}$ relative to the input image:
\begin{equation}
\mathbf{F}_i^{D} \in \mathbb{R}^{\frac{H}{8} \times \frac{W}{8} \times C_D},
\end{equation}
where $C_D$ denotes the channel dimensionality of the DINOv2 \citep{oquab2023dinov2} features. DINOv2 representations provide dense, spatially coherent embeddings that encode both semantic consistency and fine-grained structural information. Unlike task-specific optical flow encoders, these features are learned from large-scale image collections without motion supervision, and therefore capture visual regularities that generalize robustly across domains, appearance variations, and scene types. This property is particularly valuable in the zero-shot optical flow setting, where no dataset-specific adaptation is permitted.

Importantly, we keep the DINOv2 \citep{oquab2023dinov2} backbone \textbf{frozen} at train and test time. This design choice serves two purposes. First, it preserves the rich visual priors acquired during large-scale pre-training thereby preventing overfitting to motion-specific biases. Second, freezing the backbone stabilizes optimization and ensures that optical flow estimation is performed purely via inference over fixed foundation representations, rather than through task-driven representation learning. As a result, our method decouples dense correspondence estimation from optical-flow-specific feature training, thus forming a key component of our approach.

\subsection{Depth Prior Feature Encoding}
\label{ssec:depthpriorfeature}
To incorporate geometric context into dense optical flow correspondence estimation, we extract depth-aware representations from each input frame using a pretrained monocular depth foundation model. These geometric cues are particularly informative because motion discontinuities often align with depth boundaries. Concretely, given input images $\mathbf{I}_1, \mathbf{I}_2$, we compute depth features as
\begin{equation}
\mathbf{F}_i^{Z} = \Phi_{\text{Depth}}(\mathbf{I}_i),
\quad i \in \{1,2\},
\end{equation}
where $\Phi_{\text{Depth}}$ denotes the frozen Depth Anything V2-B (base) encoder \citep{yang2024depthv2}. Rather than relying on scalar depth predictions alone, we utilize intermediate feature representations produced by the depth decoder. These features encode rich geometric structure, including depth discontinuities, object boundaries, and spatial layout cues, while implicitly capturing uncertainty in regions such as occlusions and reflective surfaces \cite{jain2023enhanced} \cite{jain2024learning}. Prior works \citep{qian2025bridging, wu2025boosting} has shown that such intermediate representations often provide more informative and transferable geometric signals than final depth, particularly in downstream tasks requiring dense spatial reasoning. Since the native resolution and dimensionality of the depth features differ from those of the DINOv2 visual embeddings, we align them through a learnable projection module, i.e., 
\begin{equation}
\tilde{\mathbf{F}}_i^{Z} = \Psi_{\text{proj}}(\mathbf{F}_i^{Z}),
\end{equation}
where, $\Psi_{\text{proj}}$ is a lightweight convolutional downsampling network that maps the depth features to the same spatial resolution and channel dimensionality as $\mathbf{F}_i^{D}$. This projection enables easy integration of geometric and semantic information in subsequent processing stages, while remaining agnostic to camera intrinsics or explicit 3D reconstruction assumptions.

We emphasize that the depth encoder is kept frozen at train and test time. In doing so, we treat monocular depth estimation as a source of reusable geometric prior rather than a task to be optimized jointly with optical flow. This design ensures that our method preserves the strong generalization properties of depth foundation models and adheres to a strictly zero-shot setting, where dense optical flow is inferred entirely from fixed, pretrained representations.

\begin{figure*}[t]
\centering
\includegraphics[width=0.80\linewidth]{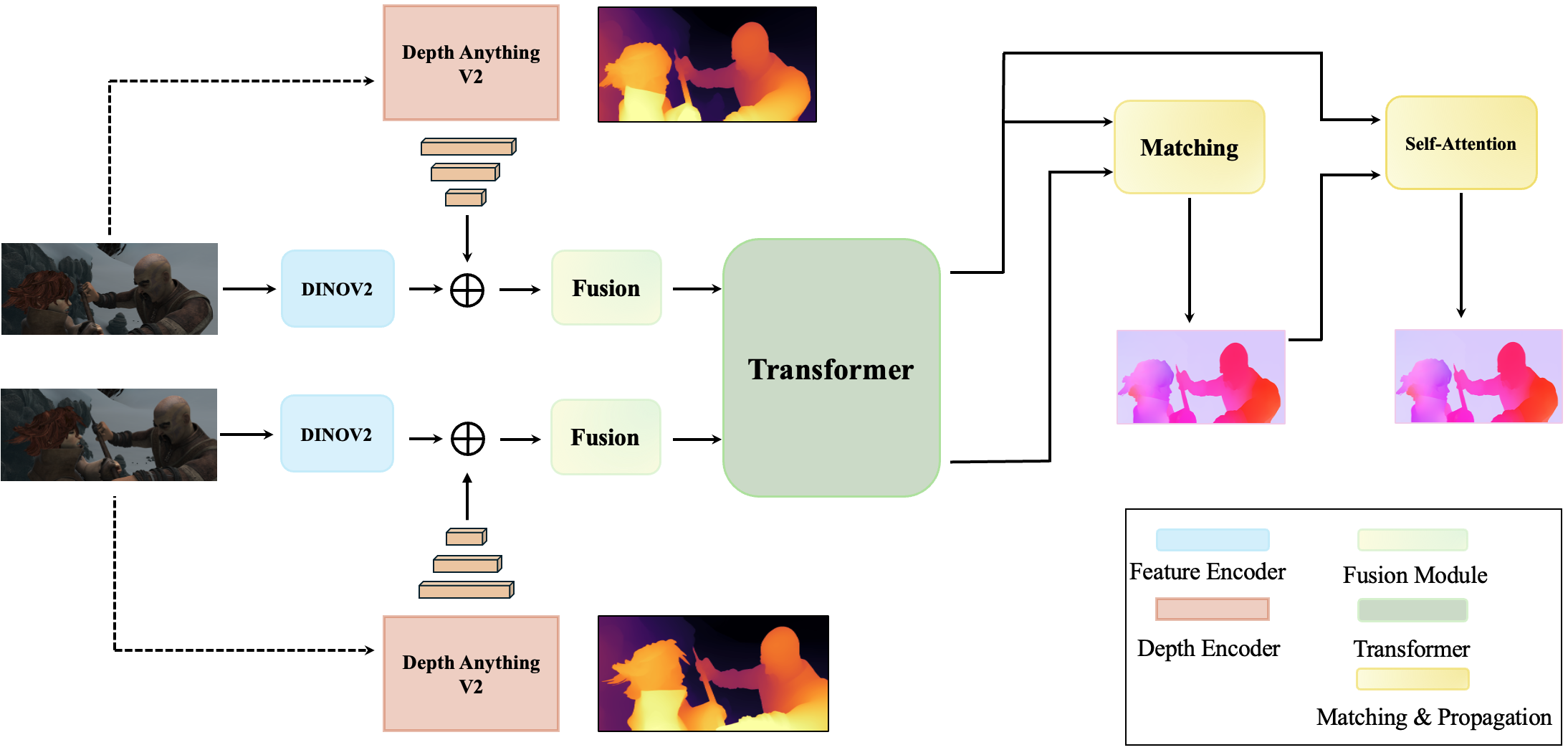}
\caption{\textbf{Overview}. The conventional CNN-based feature encoder is replaced with DINOv2 \cite{oquab2023dinov2} to provide semantically rich, large-scale self-supervised visual features, while original transformer-based feature interaction, global matching, and flow propagation modules remain unchanged. In addition, monocular depth estimates from Depth Anything V2 \cite{yang2024depthv2} are introduced as a geometric prior to improve feature conditioning and correspondence estimation.}\label{fig:flow_diagram}
\end{figure*} 

\subsection{Cross-Modal Feature Fusion}
\label{ssec:crossmodalfeature}
As alluded to above, the semantic representations extracted by DINOv2 and the geometric representations derived from the depth foundation model encode complementary, yet inherently different information. While DINOv2 features emphasize semantic consistency and visual appearance, depth features encode scene structure, boundaries, and geometric discontinuities. To effectively leverage both modalities, we construct a unified joint representation that integrates semantic and geometric cues at the feature level. Specifically, for each input frame, we first concatenate the DINOv2 features and the projected depth-aware features along the channel dimension:
\begin{equation}
\mathbf{F}_i^{C} = \text{Concat}\left(\mathbf{F}_i^{D}, \tilde{\mathbf{F}}_i^{Z}\right),
\end{equation}
where, $\mathbf{F}_i^{D}$ denotes the semantic visual features extracted by the DINOv2 backbone, and $\tilde{\mathbf{F}}_i^{Z}$ denotes the depth-aware features aligned in resolution and dimensionality as described in Sec.~\ref{ssec:depthpriorfeature}. Rather than relying on direct concatenation alone, we pass the combined representation through a learnable cross-modal fusion network $\Psi_{\text{fusion}}$. Mathematically,
\begin{equation}
\hat{\mathbf{F}}_i = \Psi_{\text{fusion}}\left(\mathbf{F}_i^{C}\right),
\end{equation}
where, $\Psi_{\text{fusion}}$ consists of lightweight convolutional layers with residual connections. This fusion network is designed to explicitly model interactions between semantic appearance cues and geometric structure. This enables the network to reweight, suppress, or reinforce features across modalities in a data-driven manner.

Our framework differs from FlowSeek \citep{Poggi_2025_ICCV}, where depth features are injected into a recurrent refinement rather than shaping the correspondence representation itself. Crucially, this fusion is performed before any correspondence matching or motion estimation. By integrating semantic and geometric information early, the resulting representation encodes both appearance similarity and structural consistency, which is essential for resolving ambiguities in challenging regions such as low-texture surfaces, repetitive patterns, motion boundaries, and illumination changes. Unlike approaches that inject geometric priors at later stages or through iterative refinement, our fusion strategy produces a single, coherent representation that the subsequent global matching and propagation pipeline can directly consume.

Finally, we emphasize that the fusion network operates entirely on frozen foundation features and introduces only a modest number of learnable parameters. This design preserves the strong cross-dataset generalization of pretrained foundation representations while allowing sufficient flexibility to reconcile modality-specific biases, ensuring that dense optical flow estimation remains an inference problem over fixed, pretrained representations rather than a task requiring domain-specific feature learning.

\subsection{Global Matching and Propagation}
\label{ssec:transformerbasedglobalmatching}
The fused feature representations $\hat{\mathbf{F}}_1, \hat{\mathbf{F}}_2$ are first processed through a transformer encoder to produce $\mathbf{F}_1$ and $\mathbf{F}_2$ $\in 
\mathbb{R}^{\frac{H}{8} \times \frac{W}{8} \times D}$ for the two input frames, we then estimate dense optical flow via a transformer-inspired global matching and propagation pipeline. This design follows the global matching formulation introduced in prior work \citep{xu2022gmflow}, but is employed here strictly as a single-pass inference mechanism operating on fixed foundation representations, without iterative refinement or test-time optimization.

\smallskip
\formattedparagraph{Global Matching.}
We first compute an all-pairs correlation volume that measures the similarity between every spatial location in the $1^\textrm{st}$ frame and every location in the second frame:
\begin{equation}
\mathbf{C}_{\text{flow}} 
= \frac{\mathbf{F}_1 \mathbf{F}_2^{\top}}{\sqrt{D}}
\in \mathbb{R}^{\frac{H}{8} \times \frac{W}{8} \times \frac{H}{8} \times \frac{W}{8}}, 
\end{equation}
where, $D$ denotes the feature dimensionality. We convert the correlation scores into a probabilistic matching distribution via a softmax over all candidate correspondences:
\begin{equation}
\mathbf{M}_{\text{flow}} = \text{softmax}\left(\mathbf{C}_{\text{flow}}\right)
\in \mathbb{R}^{\frac{H}{8} \times \frac{W}{8} \times \frac{H}{8}\times \frac{W}{8}}.
\end{equation}

Let $\mathbf{G}_{2\mathrm{D}} \in \mathbb{R}^{\frac{H}{8} \times \frac{W}{8} \times 2}$ denote the 2D coordinate grid of the second frame. The expected correspondence for each pixel in the first frame is then obtained as the expectation under the matching distribution
\begin{equation}
\hat{\mathbf{G}}_{2\mathrm{D}} = \mathbf{M}_{\text{flow}} \, \mathbf{G}_{2\mathrm{D}}
\in \mathbb{R}^{\frac{H}{8} \times \frac{W}{8} \times 2}.
\end{equation}
The initial dense optical flow field is computed as the displacement between corresponding coordinates
\begin{equation}
\hat{\mathbf{V}}_{\text{flow}} = \hat{\mathbf{G}}_{2\mathrm{D}} - \mathbf{G}_{2\mathrm{D}}
\in \mathbb{R}^{\frac{H}{8} \times \frac{W}{8} \times 2}.
\end{equation}
This formulation yields sub-pixel accurate correspondences and enables global reasoning over large displacements without relying on local search windows or recurrent updates. We intentionally keep the matching operator unchanged to isolate the effect of foundation-driven representations.

\begin{table*}[t]
\centering
\small
\setlength{\tabcolsep}{4pt}
\resizebox{1.0\linewidth}{!}{
\begin{tabular}{l|c|c|cccc|cccc}
\hline
\multirow{2}{*}{Method} & \multirow{2}{*}{\#refine} 
& \multicolumn{1}{c|}{Things (val, clean)} 
& \multicolumn{4}{c|}{Sintel (train, clean)} 
& \multicolumn{4}{c}{Sintel (train, final)} \\
\cline{3-11}
& & EPE
& EPE & s$_{0-10}$ & s$_{10-40}$ & s$_{40+}$ 
& EPE & s$_{0-10}$ & s$_{10-40}$ & s$_{40+}$ \\
\hline

RAFT \citep{teed2020raft} & 32 & 4.25 
& 1.43 & 0.33 & 1.54 & 9.04 
& 2.71 & 0.51 & 2.98 & 17.62 \\
\hline

GMFlow \citep{xu2022gmflow} & 0 & 3.48 
& 1.50 & 0.46 & 1.77 & 8.26 
& 2.96 & 0.72 & 3.45 & 17.70 \\
\hline

GMFlow \citep{xu2022gmflow} & 1 & 2.80 
& 1.08 & 0.30 & 1.25 & 6.26 
& 2.48 & 0.51 & 2.81 & 15.76 \\
\hline

SEA-RAFT (S) \citep{wang2024sea} & 4 & - 
& 1.27 & - & - & - 
& 4.32 & - & - & - \\
\hline

FlowSeek (T) \citep{Poggi_2025_ICCV} & 4 & 3.94
& 1.16 & 0.25 & 1.31 & 7.26 
& 2.48 & 0.43 & 2.63 & 16.60 \\
\hline

\textbf{Ours} & \textbf{0}
& \textbf{3.02} 
& \textbf{1.46} & 0.39 & 2.05 & 7.74 
& \textbf{2.81} & 0.64 & 3.36 & 16.99 \\
\hline
\end{tabular}
}
\caption{\small \textbf{Cross-dataset generalization} after training on Chairs and Things. No target-domain fine-tuning is applied. Lower is better.}
\label{tab:zeroshot_full}
\end{table*}

\smallskip
\formattedparagraph{Flow Propagation.}
This step effectively spreads reliable optical flow estimates across semantically and geometrically consistent regions. The softmax-based matching formulation assumes that reliable correspondences exist for all pixels, which may not hold in occluded, textureless, or out-of-boundary regions. To address this, we propagate flow estimates using feature self-similarity within the first frame. Specifically, we compute an attention matrix based on intra-frame feature affinity as
\begin{equation}
\mathbf{A} = \text{softmax}\left(
\frac{\mathbf{F}_1 \mathbf{F}_1^{\top}}{\sqrt{D}}
\right)
\in \mathbb{R}^{\frac{H}{8} \times \frac{W}{8} \times \frac{H}{8} \times \frac{W}{8}},
\end{equation}
which captures structural similarity between pixels in the reference frame. The final optical flow is obtained by propagating reliable flow estimates across similar features:
\begin{equation}
\mathbf{V} = \mathbf{A} \, \hat{\mathbf{V}}_{\text{flow}}
\in \mathbb{R}^{\frac{H}{8} \times \frac{W}{8} \times 2}.
\end{equation}
This proposed propagation allows information from confidently matched regions to inform ambiguous or occluded areas, leveraging the structural coherence encoded in the fused foundation features.

Importantly, overall matching and propagation pipeline is executed in a single forward pass, without recurrence, iterative refinement, or test-time optimization. Combined with the frozen foundation representations described in the previous sections, this design ensures that dense optical flow estimation is performed purely as inference over pretrained semantic and geometric priors, aligning with the zero-shot formulation of our approach. Figure \ref{fig:flow_diagram} provides the overall pipeline of the proposed framework.

\subsection{Training Loss}
\label{ssec:trainingloss}
We supervise flow predictions using an $\ell_1$ regression loss between the predicted flow and the ground-truth flow field. The $\ell_1$ loss provides robustness to outliers and aligns with the endpoint error metric used during evaluation. The loss $L$ is applied to both intermediate and final flow predictions, with higher weight assigned to the final prediction.
\begin{equation}
L = \sum_{i=1}^{N} \gamma^{N-i}
\left\| \mathbf{v}^{(i)} - \mathbf{v}_{gt} \right\|_{1},
\end{equation}
where, $N$ denotes the total number of flow predictions, $\mathbf{v}^{(i)}$ denotes the predicted flow at stage $i$, and $\gamma$ controls the relative weighting between intermediate and final predictions.

%% file: sec/4_experiments.tex
\section{Experiments}

\formattedparagraph{Implementation details.}
We implement our approach in PyTorch \citep{paszke2019pytorch}. Training is conducted on two NVIDIA RTX 6000 GPUs using the AdamW optimizer \citep{loshchilov2017decoupled}. Unless otherwise stated, we keep all hyperparameters fixed across experiments. We freeze DINOv2 visual backbone and the Depth Anything V2 depth backbone throughout training and inference. Only the lightweight projection, fusion, and matching modules are optimized. This design keeps the number of trainable parameters small while isolating the contributions of the pretrained semantic and geometric priors.

\smallskip
\formattedparagraph{Datasets and evaluation setup.}
We follow the standard optical flow training and evaluation protocol established in \citep{teed2020raft, wang2024sea}. We train on synthetic datasets and evaluate generalization on real-world benchmarks without target-domain fine-tuning. Specifically, we train on FlyingChairs \citep{dosovitskiy2015flownet} and FlyingThings3D \citep{mayer2016large}, and then evaluate cross-dataset generalization on the Sintel \citep{Butler:ECCV:2012} and KITTI \citep{Menze2015CVPR} training splits. For completeness and comparison with prior SOTA methods, we further report results after domain-specific fine-tuning on Sintel and KITTI 2015. These fine-tuned results are presented separately and do not support our main claims about generalization without target-domain adaptation.

\smallskip
\formattedparagraph{Metrics.}
We evaluate optical flow accuracy using the standard endpoint error
(EPE), defined as the average $\ell_2$ distance between predicted and ground-truth flow vectors over all pixels. To provide a more fine-grained analysis across motion regimes, we report EPE stratified by ground-truth flow magnitude: $s_{0\text{--}10}$, $s_{10\text{--}40}$, and $s_{40+}$, corresponding to pixel displacements of $0$--$10$, $10$--$40$, and greater than $40$ pixels, respectively. For Sintel, we report both overall EPE and these motion-stratified metrics. For KITTI 2015, we also report F1-all, which measures the percentage of outlier pixels under the official KITTI evaluation protocol.

\smallskip
\formattedparagraph{Training protocol.}
Training proceeds in stages. We first train on FlyingChairs for 200k iterations using a batch size of 16, a learning rate of $4 \times 10^{-4}$, and random crops of size $384 \times 512$. We then continue training on FlyingThings3D for 800k iterations with a reduced learning rate of $2 \times 10^{-4}$ and crop size $384 \times 768$. For fine-tuning experiments, we adopt a mixed training set consisting of KITTI \citep{geiger2013vision}, HD1K \citep{kondermann2016hci}, FlyingThings3D, and Sintel (denoted as TSKH), and train for an additional 200k iterations using crops of size $320 \times 896$. Finally, for the KITTI 2015 evaluation, we fine-tune the model for 90k iterations with a batch size of 8, a learning rate of $2 \times 10^{-4}$, and a crop size of $320 \times 1152$.

\subsection{Cross-Dataset Generalization}
\label{ssec:cross-dataset}
Table~\ref{tab:zeroshot_full} summarizes cross-dataset generalization results after training exclusively on FlyingChairs and FlyingThings3D, with no target-domain fine-tuning at test time. This setting is deliberately challenging, as it probes whether a model trained on synthetic data can transfer to domains with substantially different appearance, motion statistics, and rendering characteristics. Overall, our method generalizes strongly across synthetic and real benchmarks. On the FlyingThings3D validation set, our approach achieves an EPE of 3.02, outperforming RAFT and GMFlow without refinement, while competitive with methods that rely on iterative updates. Notably, this performance is obtained without recurrent refinement or test-time optimization.

On Sintel, our method shows a clear advantage on more challenging Final pass. In particular, we achieve an EPE of \textbf{2.81} on Sintel Final, substantially improving over SEA-RAFT (4.32 EPE) despite SEA-RAFT employing multiple refinement steps. Compared to GMFlow without refinement, our method also achieves lower overall error on Sintel Final while operating in a strictly single-pass inference regime. When compared to FlowSeek \citep{Poggi_2025_ICCV}, our approach attains comparable performance on both Sintel passes, despite FlowSeek benefiting from additional pretraining on TartanAir. Taken together, these results support our central hypothesis: stronger semantic and geometric priors can partially substitute for increased inference-time computation in dense optical flow estimation.

\subsection{Results on Sintel and KITTI}
\label{ssec:results}

\formattedparagraph{Results on Sintel.}
Table~\ref{tab:sintel_trainedg} reports performance on Sintel train set after training on Chairs, Things, and the mixed TSKH dataset. Overall, our method achieves competitive performance on both Clean and Final passes, despite operating in a single-pass setting without iterative refinement or additional large-scale pretraining. On the more challenging Sintel Final split, our approach outperforms RAFT, GMFlow without refinement, and FlowSeek, while remaining close to refinement-based GMFlow. This trend is notable because refinement-based methods explicitly revisit and correct flow estimates via multiple update iterations, whereas our model relies on a single forward pass over frozen foundation representations. These results shows that the use of visual semantic and geometric representation priors captures sufficient global context to handle the appearance changes and motion patterns present in Sintel Final.

SEA-RAFT \cite{wang2024sea} achieves the strongest performance on both Sintel splits. As shown in Table~\ref{tab:sintel_trainedg}, this gap likely reflects differences in both training scale and auxiliary pretraining. In particular, SEA-RAFT is trained using substantially larger compute resources, e.g., $8\times$ NVIDIA L40 GPUs compared to $2\times$ RTX 6000 GPUs in our setup, corresponding to approximately \textbf{$4\times$} higher effective training compute. This difference is also reflected in larger effective batch sizes and higher-resolution training crops (batch size 32 with $432 \times 960$ crops for SEA-RAFT versus batch size 8 with $320 \times 896$ crops in our training). Moreover, both SEA-RAFT and FlowSeek benefit from additional pretraining on TartanAir, which introduces greater scene diversity and motion variability beyond the TSKH mixture. Despite these advantages, our approach remains competitive, highlighting the effectiveness of foundation-model-driven representations for dense optical flow estimation.

\begin{table}[t]
\centering
\small
\setlength{\tabcolsep}{4pt}
\resizebox{1.0\linewidth}{!}{
\begin{tabular}{l|c|c|c|c}
\hline
Method & Extra Data & \#refine & Sintel Clean (EPE) & Sintel Final (EPE) \\
\hline
RAFT \citep{teed2020raft} 
& -- 
& 32 
& 0.768 & 1.217 \\
\hline

GMFlow \citep{xu2022gmflow} 
& -- 
& 0 
& 0.947 & 1.276 \\
\hline

GMFlow \citep{xu2022gmflow} 
& -- 
& 1 
& 0.762 & 1.110 \\
\hline

SEA-RAFT (S) \citep{wang2024sea} 
& TartanAir 
& 4 
& 0.546 & 0.782 \\
\hline

FlowSeek (T) \citep{Poggi_2025_ICCV} 
& TartanAir 
& 4 
& 0.71 & 1.28 \\
\hline

\textbf{Ours} 
& \textbf{--}
& \textbf{0}
& \textbf{0.847} & \textbf{1.140} \\
\hline
\end{tabular}
}
\caption{\textbf{Performance on the Sintel train set} after training on Chairs, Things, and mixed datasets (TSKH). Our method outperforms RAFT, GMFlow (wo refinement), and FlowSeek on Sintel Final. SEA-RAFT performs best on both splits, likely due to differences in training scale and extra pretraining, including substantially larger effective batch sizes \textbf{(32 with crop size $432 \times 960$ for SEA-RAFT versus 8 with crop size $320 \times 896$ in our training)}.}
\label{tab:sintel_trainedg}
\end{table}

\begin{figure*}[t]
\centering
\begin{subfigure}[t]{0.19\textwidth}
    \includegraphics[height=0.06\textheight,width=\linewidth]{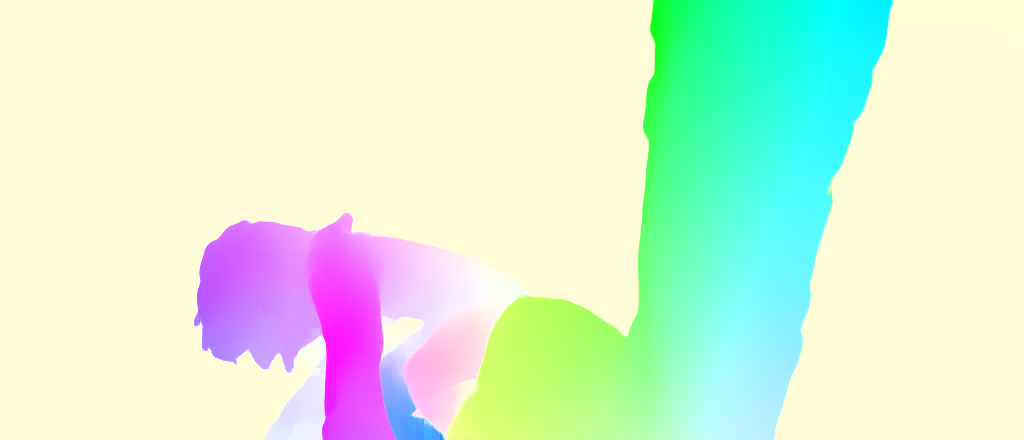}
\end{subfigure}
\begin{subfigure}[t]{0.19\textwidth}
    \includegraphics[height=0.06\textheight,width=\linewidth]{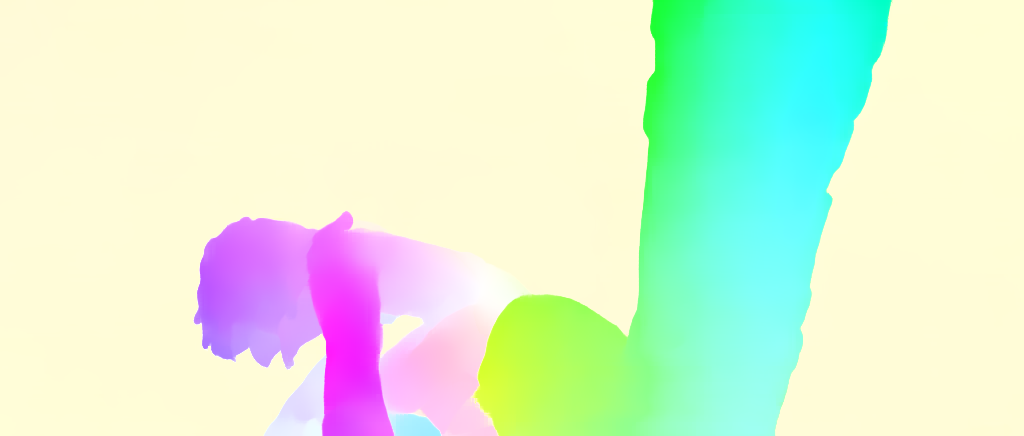}
\end{subfigure}
\begin{subfigure}[t]{0.19\textwidth}
    \includegraphics[height=0.06\textheight, width=\linewidth]{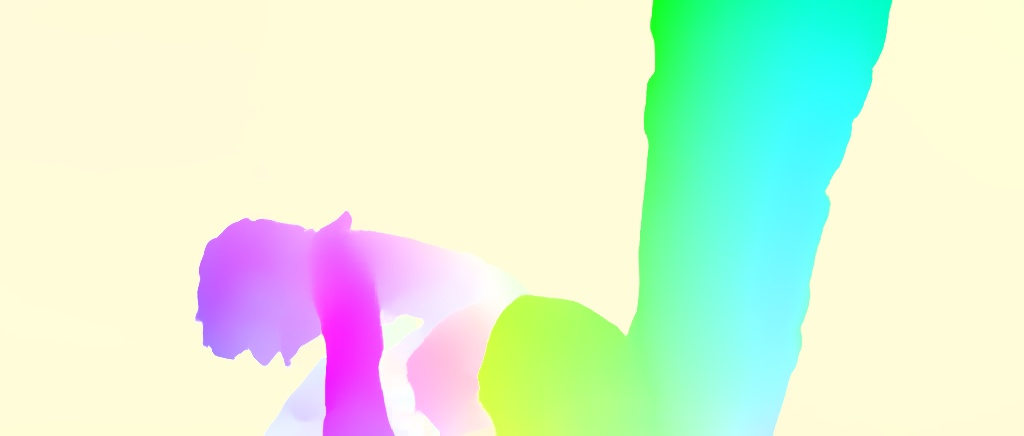}
\end{subfigure}
\begin{subfigure}[t]{0.19\textwidth}
    \includegraphics[height=0.06\textheight,width=\linewidth]{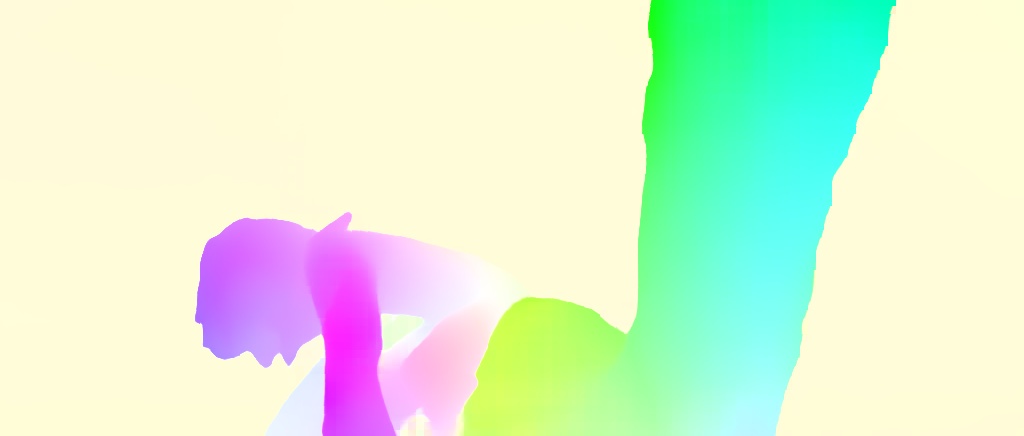}
\end{subfigure}
\begin{subfigure}[t]{0.19\textwidth}
    \includegraphics[height=0.06\textheight,width=\linewidth]{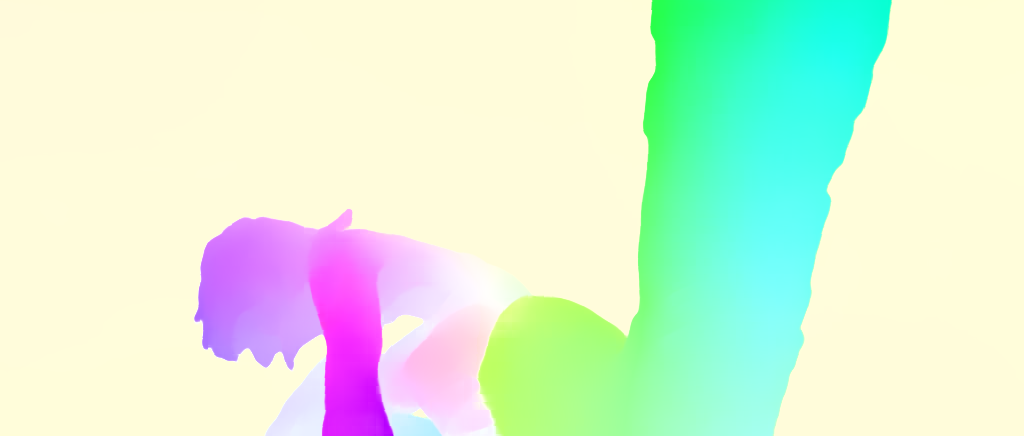}
\end{subfigure}

\begin{subfigure}[t]{0.19\textwidth}
    \includegraphics[height=0.06\textheight,width=\linewidth]{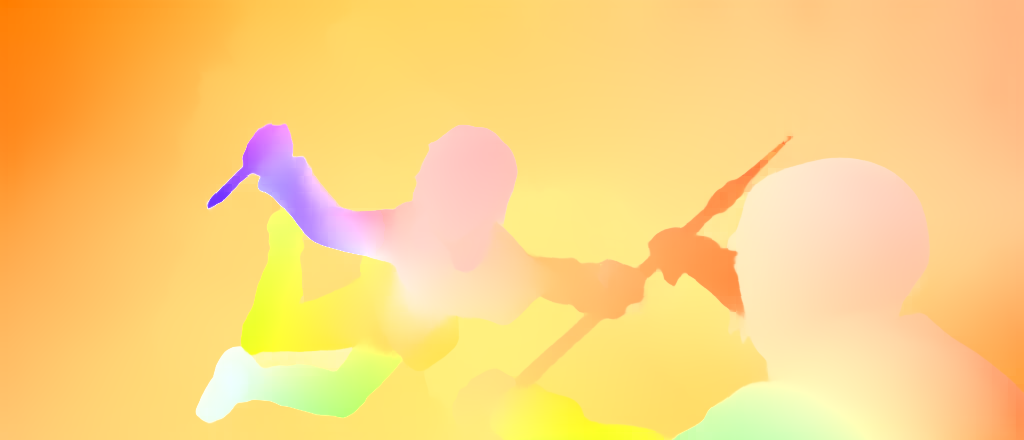}
\end{subfigure}
\begin{subfigure}[t]{0.19\textwidth}
    \includegraphics[height=0.06\textheight,width=\linewidth]{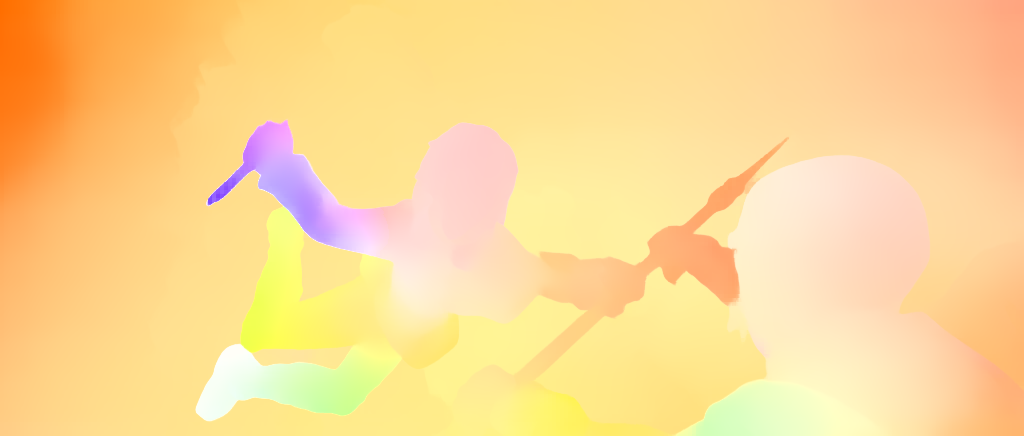}
\end{subfigure}
\begin{subfigure}[t]{0.19\textwidth}
    \includegraphics[height=0.06\textheight,width=\linewidth]{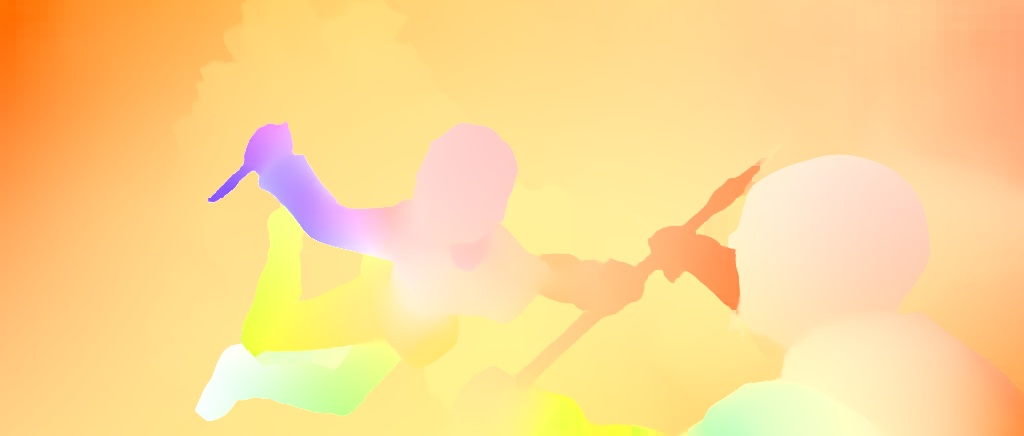}
\end{subfigure}
\begin{subfigure}[t]{0.19\textwidth}
    \includegraphics[height=0.06\textheight,width=\linewidth]{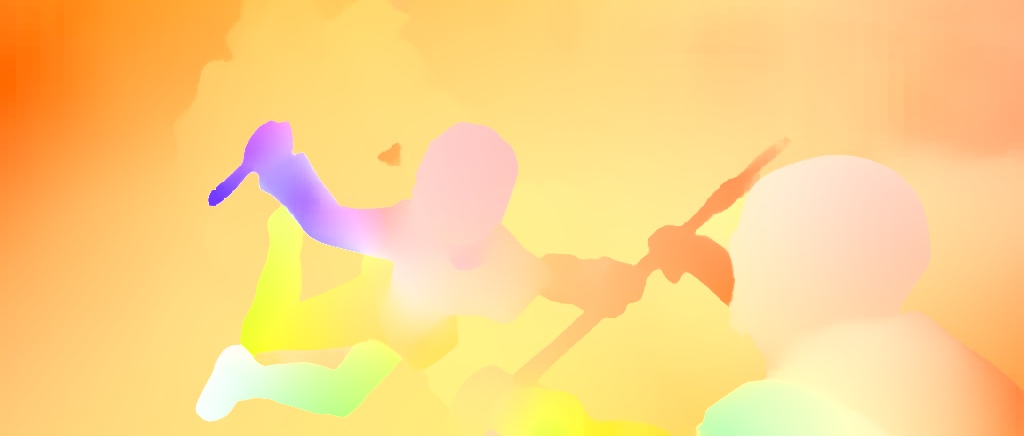}
\end{subfigure}
\begin{subfigure}[t]{0.19\textwidth}
    \includegraphics[height=0.06\textheight,width=\linewidth]{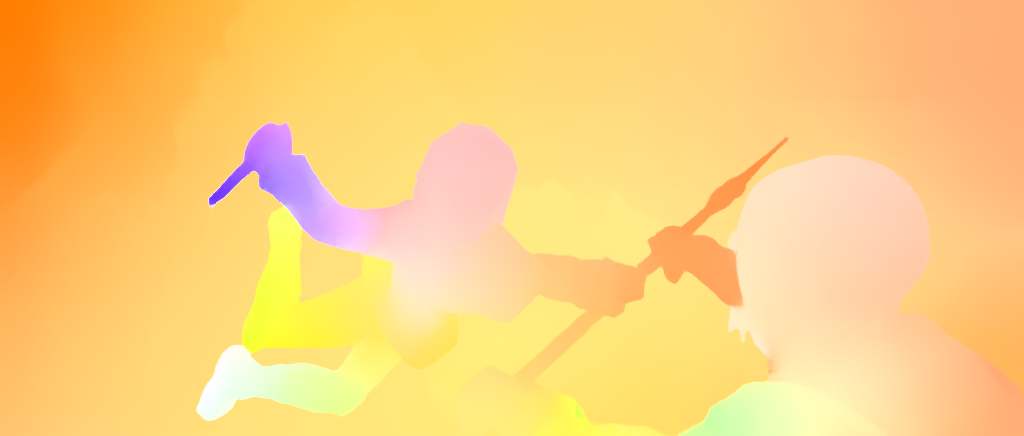}
\end{subfigure}

\vspace{3pt}

\begin{subfigure}[t]{0.19\textwidth}
    \includegraphics[height=0.06\textheight, width=\linewidth]{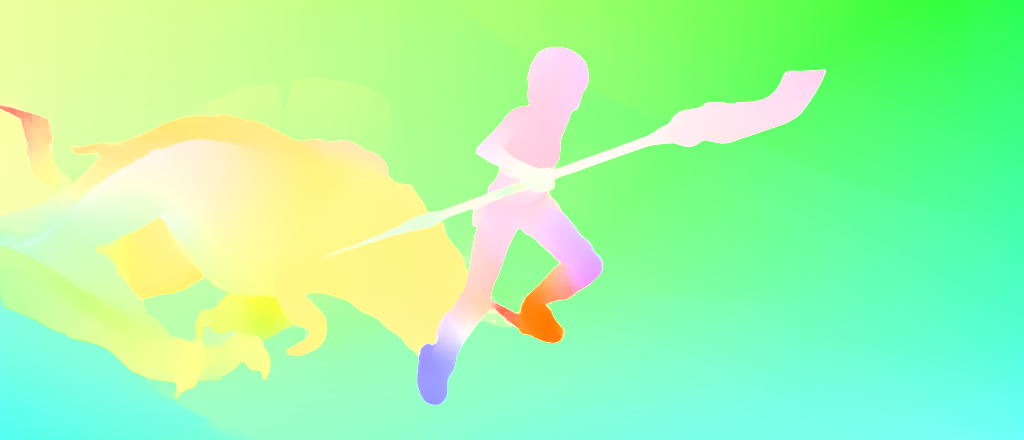}
\end{subfigure}
\begin{subfigure}[t]{0.19\textwidth}
    \includegraphics[height=0.06\textheight,width=\linewidth]{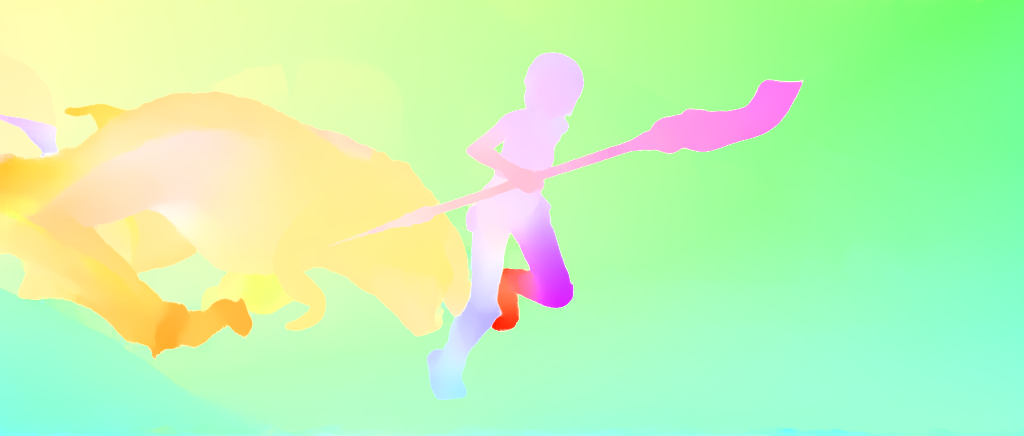}
\end{subfigure}
\begin{subfigure}[t]{0.19\textwidth}
    \includegraphics[height=0.06\textheight,width=\linewidth]{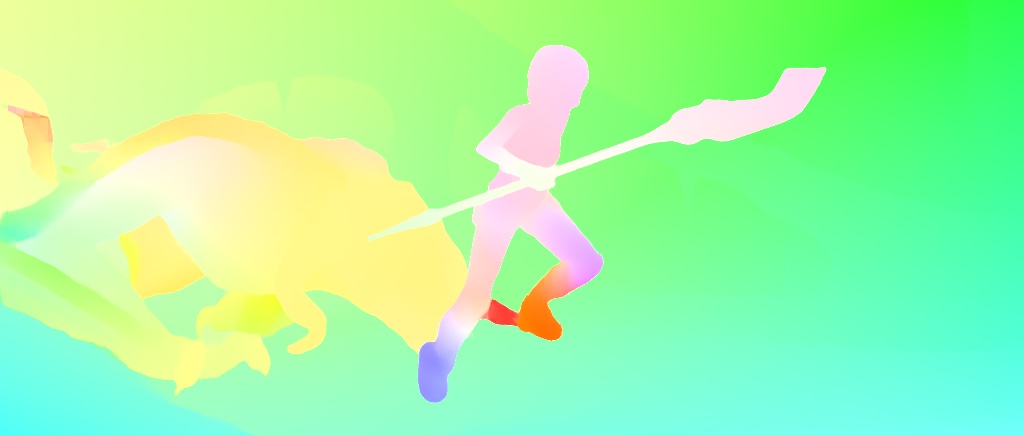}
\end{subfigure}
\begin{subfigure}[t]{0.19\textwidth}
    \includegraphics[height=0.06\textheight,width=\linewidth]{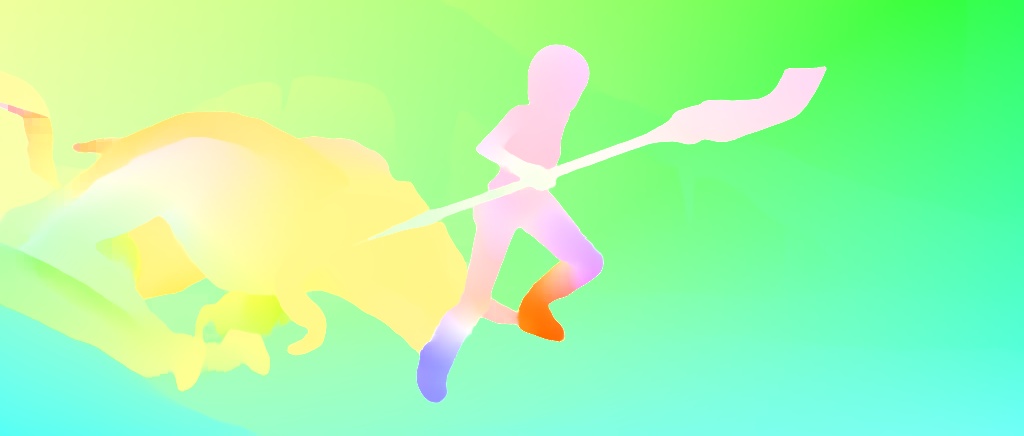}
\end{subfigure}
\begin{subfigure}[t]{0.19\textwidth}
    \includegraphics[height=0.06\textheight,width=\linewidth]{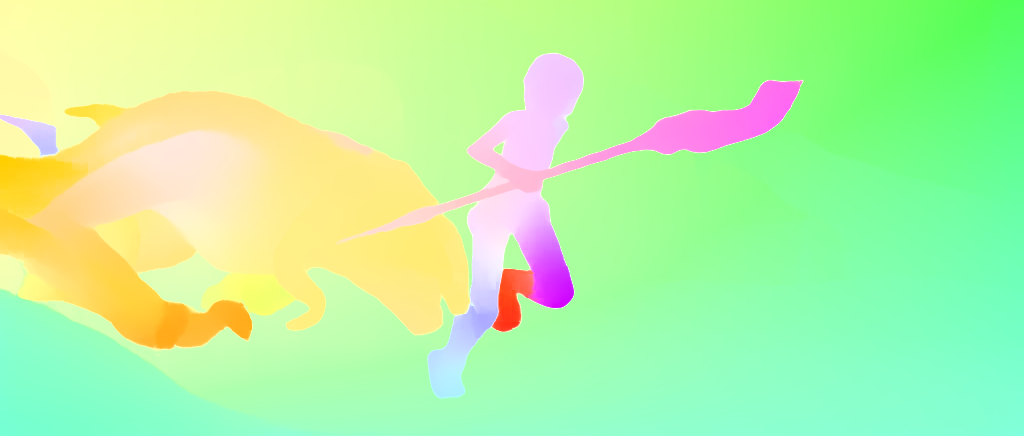}
\end{subfigure}

\vspace{3pt}

\begin{subfigure}[t]{0.19\textwidth}
    \includegraphics[height=0.06\textheight,width=\linewidth]{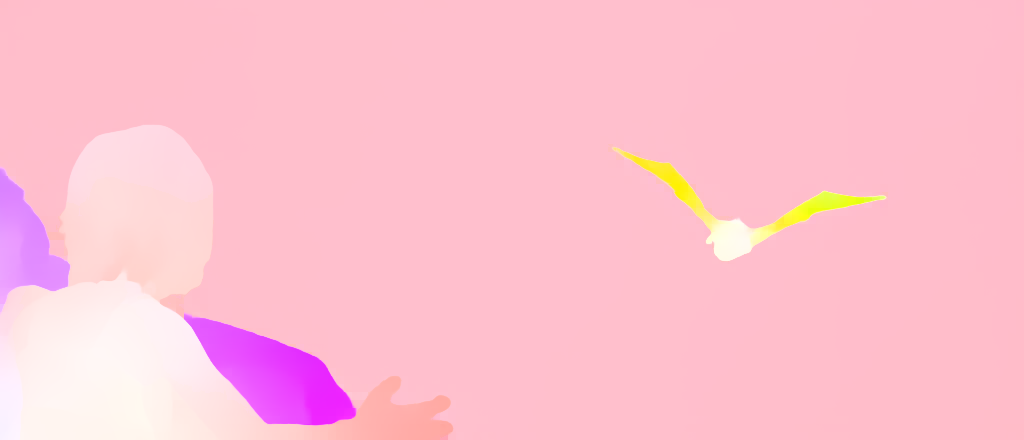}
\end{subfigure}
\begin{subfigure}[t]{0.19\textwidth}
    \includegraphics[height=0.06\textheight,width=\linewidth]{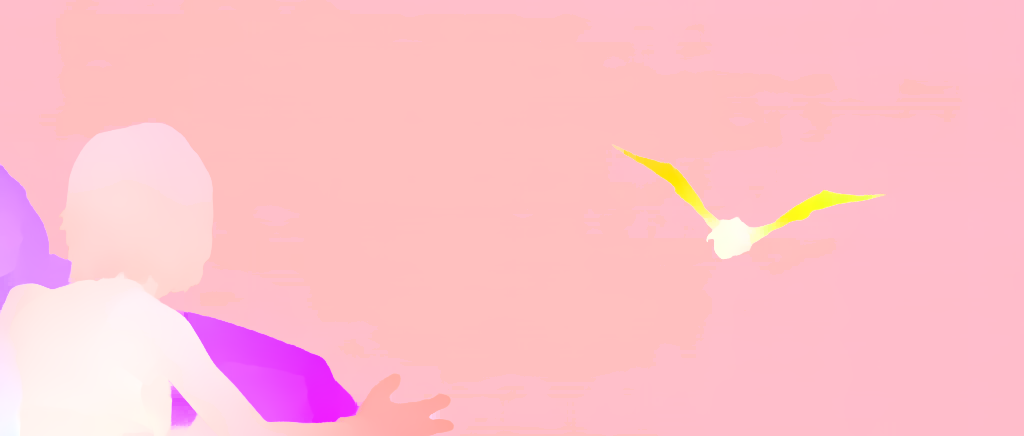}
\end{subfigure}
\begin{subfigure}[t]{0.19\textwidth}
    \includegraphics[height=0.06\textheight,width=\linewidth]{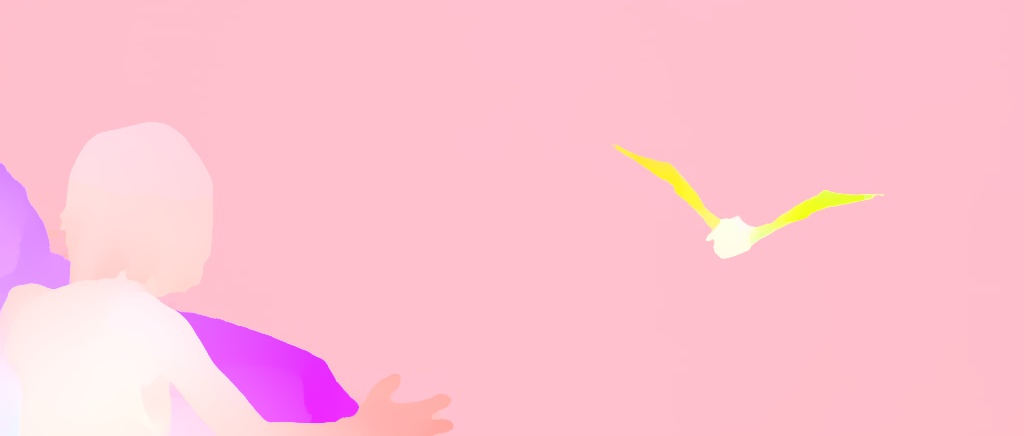}
\end{subfigure}
\begin{subfigure}[t]{0.19\textwidth}
    \includegraphics[height=0.06\textheight,width=\linewidth]{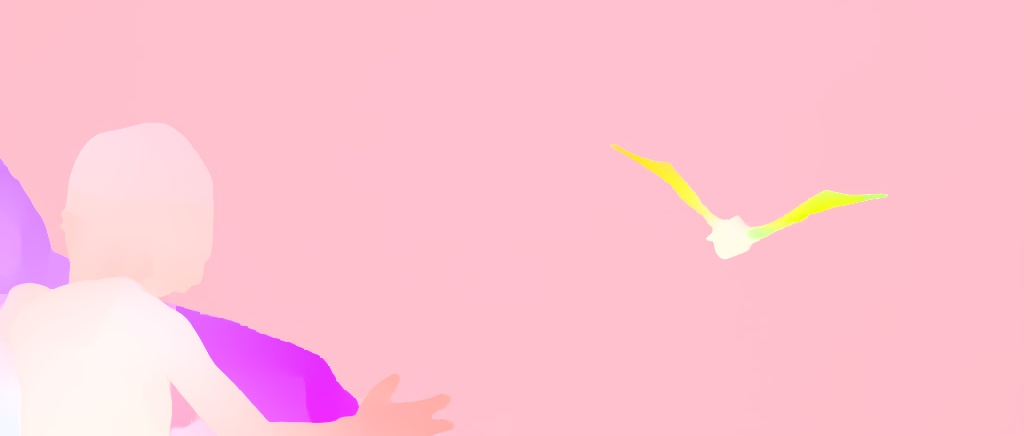}
\end{subfigure}
\begin{subfigure}[t]{0.19\textwidth}
    \includegraphics[height=0.06\textheight,width=\linewidth]{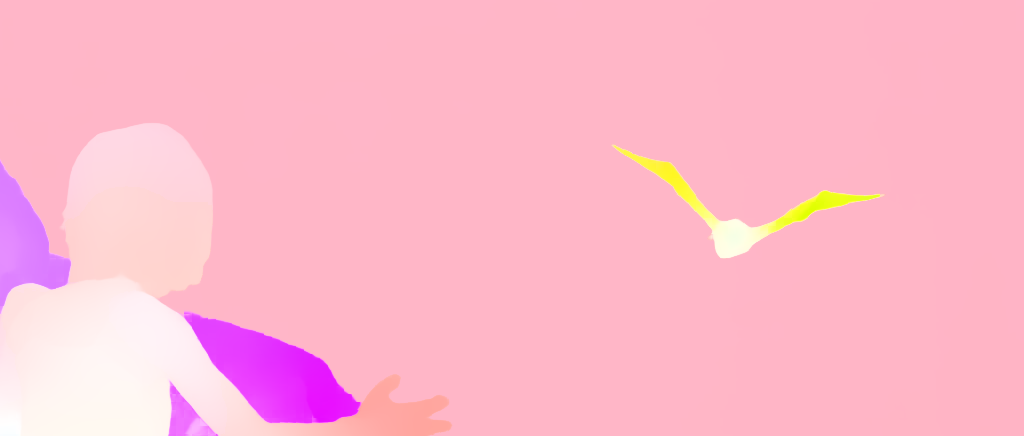}
\end{subfigure}

\vspace{2pt}

\begin{subfigure}[t]{0.19\textwidth}\centering{\small RAFT \cite{teed2020raft}}\end{subfigure}
\begin{subfigure}[t]{0.19\textwidth}\centering{\small GMFlow \cite{xu2022gmflow}}\end{subfigure}
\begin{subfigure}[t]{0.19\textwidth}\centering{\small SEA-RAFT \cite{wang2024sea}}\end{subfigure}
\begin{subfigure}[t]{0.19\textwidth}\centering{\small FlowSeek \cite{Poggi_2025_ICCV}}\end{subfigure}
\begin{subfigure}[t]{0.19\textwidth}\centering{\small \textbf{Ours}}\end{subfigure}

\vspace{2pt}
\caption{\small \textbf{Qualitative comparison on Sintel (Final).} This benchmark contains severe motion blur, illumination variation, and large displacements. Despite operating without iterative refinement, our method preserves sharp motion boundaries and produces accurate flow estimates, performing comparably to---and in some cases better than---refinement-based approaches.}
\label{fig:qual_flow}
\end{figure*}

\smallskip
\formattedparagraph{Results on KITTI.}
Table~\ref{tab:kitti_trainedg} presents results on the KITTI train set after fine-tuning on KITTI following training on Chairs, Things, and TSKH. Our framework achieves performance comparable to GMFlow without refinement, while using frozen foundation priors rather than learned flow-specific encoders. Although refinement-based methods such as SEA-RAFT and FlowSeek obtain lower EPE and F1-all scores, they benefit from both recurrent update mechanisms and pretraining on TartanAir, which provides optical flow priors and scene structures closely aligned with KITTI.

KITTI poses additional challenges due to frequent occlusions, thin structures, and sharp motion boundaries. Iterative refinement is particularly effective in such scenarios as it allows a model to correct local errors and enforce spatial consistency. In contrast, our framework deliberately avoids refinement to maintain a simple, scalable inference pipeline based on frozen foundation model priors. The observed performance gap, therefore, reflects a trade-off between architectural simplicity and fine-grained local accuracy.

For completeness, we note that FlowSeek \cite{Poggi_2025_ICCV} does not release a checkpoint trained exclusively on KITTI; consequently, we evaluate it using the publicly available model trained under a mixed-dataset setting. Overall, these results indicate that while refinement and additional pretraining remain beneficial for achieving peak performance on KITTI, foundation-model-driven representations already provide a strong baseline. Incorporating lightweight refinement mechanisms or larger-scale pretraining remains a promising direction for improving performance further without abandoning the foundation-based paradigm.

\begin{table}[h]
\centering
\small
\setlength{\tabcolsep}{4pt}
\resizebox{1.0\linewidth}{!}{
\begin{tabular}{l|c|c|c|c}
\hline
Method & Extra Data & \#refine & KITTI EPE & KITTI F1-all \\
\hline

RAFT \citep{teed2020raft} 
& -- 
& 32 
& 0.63 & 1.47 \\
\hline

GMFlow \citep{xu2022gmflow} 
& -- 
& 0 
& 2.06 & 7.57 \\
\hline

GMFlow \citep{xu2022gmflow} 
& -- 
& 1 
& 1.36 & 5.17 \\
\hline

SEA-RAFT (S) \citep{wang2024sea} 
& TartanAir 
& 4 
& 0.93 & 2.65 \\
\hline

FlowSeek (T) \citep{Poggi_2025_ICCV} 
& TartanAir 
& 4 
& 1.26 & 3.90 \\
\hline

\textbf{Ours} 
& \textbf{--}
& \textbf{0}
& 1.99 & 7.40 \\
\hline
\end{tabular}
}
\caption{\textbf{Performance on the KITTI train set} after fine-tuning on KITTI, following training on Chairs, Things, and the mixed TSKH dataset.}
\label{tab:kitti_trainedg}
\end{table}

\section{Ablation}\label{sec:ablation}
\formattedparagraph{1. Performance analysis of the proposed fusion module.}
We analyzed the impact of the proposed fusion module, which integrates visual semantic features from DINOv2 with geometric cues from the depth foundation model. As shown in Table~\ref{tab:fusion_ablation}, enabling the fusion module consistently improves cross-dataset generalization across popular benchmarks. On FlyingThings3D, fusion reduces the EPE from 3.52 to 3.02, indicating that combining visual and geometric priors improves correspondence estimation even on synthetic validation data. The improvement is more pronounced on the Sintel benchmark. On the Clean split, fusion reduces EPE from 1.575 to 1.46, while on the more challenging Final split, the error decreases from 3.12 to 2.81, corresponding to approximately a 10\% relative improvement. A motion-stratified study further highlights the benefits of fusion across displacement regimes. In particular, the largest gain appears in the large-motion regime ($s_{40+}$) on Sintel Final, where EPE decreases from 19.37 to 16.99.

\begin{table}[h]
\centering
\resizebox{\linewidth}{!}{
\begin{tabular}{l|c|cccc|cccc}
\hline
\multirow{2}{*}{Method}
& \multicolumn{1}{c|}{Things (val, clean)} 
& \multicolumn{4}{c|}{Sintel (train, clean)} 
& \multicolumn{4}{c}{Sintel (train, final)} \\
\cline{2-10}
& EPE
& EPE & s$_{0-10}$ & s$_{10-40}$ & s$_{40+}$ 
& EPE & s$_{0-10}$ & s$_{10-40}$ & s$_{40+}$ \\
\hline

(w/o) Fusion & 3.52
& 1.575 & 0.45 & 2.01 & 8.53
& 3.12 & 0.685 & 3.57 & 19.37 \\

(w/) Fusion & 3.02
& 1.46 & 0.39 & 2.05 & 7.74
& 2.81 & 0.64 & 3.36 & 16.99 \\
\hline
\end{tabular}
}
\caption{\small Ablation study evaluating the contribution of the proposed fusion module.}\label{tab:fusion_ablation}
\end{table}

\formattedparagraph{2. Usefulness of depth features.} To analyse the usefulness of depth feature, we train the depth model for 100k iterations on the FlyingChairs dataset with and without depth features. For evaluating the variant without depth, we also exclude the cross-modal fusion module used for feature integration. We observed that incorporating depth features leads to a significant improvement in dense optical flow estimation, reducing the EPE from \textbf{1.77} to \textbf{0.87}.

%% file: sec/5_conclusion.tex
\section{Conclusion and Limitations}
In this paper, we present a dense optical flow estimation framework that utilizes pretrained complementary visual priors from foundation models. By using frozen representations from DINOv2 \cite{oquab2023dinov2} and a monocular depth from \cite{yang2024depthv2} foundation model and integrating them via a lightweight fusion and global matching pipeline, our approach estimates optical flow in a single forward pass without task-specific backbone training, iterative refinement, or test-time optimization. Experimental evaluations show strong cross-dataset generalization. Moreover, our ablation show that strong foundation model representations have the potential to substitute for test-time scaling in dense optical flow.

Despite encouraging results, our work has a few limitations. First, our proposed approach may yield lower accuracy in challenging scenarios involving heavy occlusions, thin structures, or fine motion boundaries. Next, our approach depends on the availability and quality of large pretrained foundation models, whose training costs lie outside the scope of this work and whose representations may carry biases inherited from large-scale datasets. Addressing these limitations opens promising research directions.

\formattedparagraph{Acknowledgment.} The authors thank High Performance Research Computing (HPRC) at Texas A\&M University, Texas, USA for providing us with the startup credits for utilizing the GPU-server facility. 